\documentclass[review]{elsarticle}

\usepackage{subcaption}
\usepackage[hidelinks]{hyperref}
\usepackage{float}
\usepackage{algpseudocode}
\usepackage[tableposition=top]{caption}
\usepackage{amsmath}
\usepackage[ruled,linesnumbered]{algorithm2e}
\usepackage{setspace}
\journal{Preprint}
\begin{document}
\begin{frontmatter}
\title{An improved clustering-based multi-swarm PSO using local diversification and topology information}

% \author[yves.matanga@gmail.com]
% {Yves Matanga\fnref{myfootnote}}\cortext[mycorrespondingauthor]{}
% \ead{yves.matanga@gmail.com}
% \ead[url]{www.uj.ac.za}

% \author[ysun@uj.ac.za]{Yanxia Sun}
% \ead[url]{www.uj.ac.za}

% \author[KurienAM@tut.ac.za]{Anish Kurien}
% \ead[url]{www.tut.ac.za}
% \address{Department of Electrical and Electronic Engineering Science, University of Johannesburg, Johannesburg 2006, South Africa}

\author[First]{Yves Matanga \corref{cor1}\fnref{label2}}
\ead{yves.matanga@gmail.com}

\author[Second]{Yanxia Sun}
\ead{ysun@uj.ac.za}

\author[Third]{Zenghui Wang}
\ead{wangz@unisa.ac.za}
% \fntext[label2]{}
\cortext[cor1]{Corresponding Author}
\address[First]{Department of Electrical and Electronic Engineering Science, University of Johannesburg, Johannesburg 2006, South Africa }
\address[Second]{Department of Electrical and Electronic Engineering Science, University of Johannesburg, Johannesburg 2006, South Africa}
\address[Third]{Department of Electrical and Mining Engineering, University of South Africa, Johannesburg 1710, South Africa}

\begin{abstract}
Multi-swarm particle optimisation algorithms are gaining popularity due to their ability to locate multiple optimum points concurrently. In this family of algorithms, clustering-based multi-swarm algorithms are among the most effective techniques that join the closest particles together to form independent niche swarms that exploit potential promising regions. However, most clustering-based multi-swarms are euclidean distance based and only inquire about the potential of one peak within a cluster and thus can lose multiple peaks due to poor resolution. In a bid to improve the peak detection ratio, the current study proposes two enhancements. First, a preliminary local search across initial particles is proposed to ensure that each local region is sufficiently scouted prior to particle collaboration. Secondly, an investigative clustering approach that performs concavity analysis is proposed to evaluate the potential for several sub-niches within a single cluster. An improved clustering-based multi-swarm PSO 
 (TImPSO) has resulted from these enhancements and tested against three competing algorithms in the same family using the IEEE CEC2013 niching datasets, resulting in an improved peak ratio for almost all the test functions.
\end{abstract}

\begin{keyword}
\texttt{particle swarm optimisation\sep geometric topology \sep clustering \sep sub-clustering \sep local diversification}
\end{keyword}
\end{frontmatter}
% \linenumbers

\section{Introduction}
\noindent Multiswarm PSO techniques have been proposed throughout the years to mitigate the unimodal search attitude of classical PSO. Unlike classical PSO, multiswarm PSO algorithms discover several promising areas independently and propose multiple global optima at the end of the search mechanism. These algorithms are typically called niching strategies, within which several paradigms exist that can be subdivided into two large families: parallel niching or sequential niching approaches \cite{beasley1993sequential}. While sequentially niching algorithms search for promising basins of convergence, one at a time, parallel niching algorithms use particles independently, each heading concurrently towards different promising regions. This approach has gained popularity in literature and is generally argued to be faster. In the family of parallel niching strategies, Clustering-based parallel niching algorithms present good scalability properties and thus deserve continued attention \cite{kronfeld2010towards}. Clustering-based multi-swarm algorithms typically proceed in a three-stage approach: preliminary search, niche formation and fine search. Initially, each particle navigates the problem space independently or using a loosely coupled neighbourhood topology. When the particles' cognitive experiences are no longer significantly enriched, clustering is performed to join closely spaced particles into niche swarms, which make use of the gBest topology or similar variants to improve the quality of the solutions found in each niche \cite{passaro2008particle,liu2020niching,streichert2003clustering}. This approach has yielded good results in niching test benches mostly evaluated with problems with a limited number of optima. The current research study has pinpointed two shortcomings of clustering-based multi-swarm PSO algorithms. First, during the preliminary search, particles do not sufficiently explore their local regions. The use of the cognitive model, for instance, drives particles to navigate back and forth around their best experience so far, which does not guarantee pervasive exploration. Small neighbour topologies, on the other hand, by design, tend to favour the exploration of the best region among neighbours, also hindering local exploration, thus defeating the purpose of the preliminary search (i.e., maximal regional information harvesting). In light of the above, the current research has proposed a preliminary scouting phase around each particle region which should be uniformly distributed in the problem space, followed by a cognitive search to enrich the estimate of the scouting phase. This model is referred to as the pervasive-cognitive model. The second and most important limitation of clustering-based multi-swarm PSO (CMPSO) algorithms is the poor resolution of the clustering features used. CMPSO algorithms cluster particles into a single niche based on proximity without investigating the possible existence of sub-niches within a single cluster. Neighbouring particles do not necessarily belong to the peak. This approach on its own can lead to losing several peaks as every single niche fine-tunes a single optimum. The current research proposes that a further investigation of the formed clusters be performed using geometric topology information to identify potential sub-niches and thus increase the peak detection ratio. These two additional enhancements have been integrated to conventional CMPSO algorithms yielding a higher discovery rate of existing niches when tested on IEEE CEC2013 niching datasets. The main contributions of this paper are:
\begin{enumerate}
    \item A preliminary scouting model is proposed to augment the local information capacity of exploration particles with an impact on accurate niche formation.
    \item A sub-clustering algorithm is proposed to identify sub-niches within preliminary clusters using concavity analysis which improves the peak ratio performance of CMPSO algorithms
    \item A comparative analysis of results is performed with three CMPSO algorithms ascertaining the relevance of the proposed enhancements.    
\end{enumerate}
\noindent

\noindent The rest of the paper is subdivided as follows: Section 2 gives a brief description of the particle swarm algorithm. Section 3 discusses existing CMPSO variants, including NichePSO, a 
 representative multiswarm PSO algorithm. Section 4 presents the proposed algorithmic enhancements. Section 5 describes the computation experiment used to assess the contribution of the enhancements. Section 6 presents and discusses the results, and Section 7 gives a conclusion to the research study. 
\section{Classical particle swarm optimisation}
\noindent Particle swarm optimisation (PSO) is a population-based optimisation algorithm inspired by the foraging of flocks of birds or insects in a collaborative methodology \cite{kennedy1995particle}. A set of randomly generated potential solutions is generated, and each individual (particle) iteratively improves its position based on its own experience (cognitive learning) and based on the experience of other individuals in the swarm (social learning). The motion of each particle in the problem space is thus dictated by a collaborative randomised search direction:

\begin{eqnarray}
    v_i^{k+1}=\omega^kv_i^{k}+ c_1r_1^k(p_i^k-x_i^k)+c_2r_2^k(g_{(i)}^k-x_i^k)\\
    x_i^{k+1} = x_i^k + v_i^{k+1}
\end{eqnarray}

\noindent where $v_i^{k+1}$ is the search direction of a given particle for the next iteration, a function of its current velocity $v_i^k$, the location of its best location thus far \textbf{$p_i^k$} (cognitive learning) and the location of the global best location $g_{(i)}^k$ (social learning) in the neighbourhood of the particle ($g_i^k$) or within the whole swarm ($g^k$). When $g_{(i)}^k$ represents the best candidate in a neighbourhood, the algorithm is referred to as local and likewise global when $g_{(i)}^k$ represent the best candidate in the whole population \cite{bonyadi2017particle}. $c_1$ and $c_2$ represent acceleration parameters for the cognitive and social learning components. $r_1^k,r_2^k\in [0,1]$ are uniform randomly generated numbers that simulate the stochastic behaviour of the swarm. 

\begin{figure}[h]
    \centering
    \includegraphics[scale=0.3]{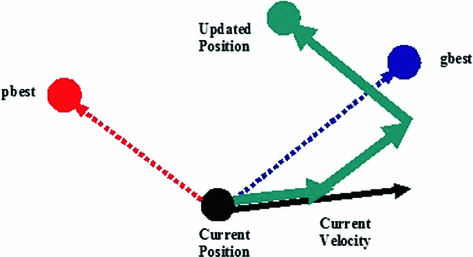}
    \caption{Spatial illustration of a particle movement in PSO \cite{Bansal2019}}
    \label{fig:my_label}
\end{figure}

The original publication proposes the value of $c_1=c_2=2$, and these settings are mainly used and accepted in the literature \cite{rezaee2013parameter}. The inertial parameter $w^k$ defines how willing a given particle is to maintain its current direction. It contributes to dictating bias towards exploration and exploitation. The higher the inertial parameter, the more exploratory the search. In the original publication, the inertial parameter was set to 1 ($w^k=w=1$). A more adaptive variation $w^k$ is used in recent literature \cite{Mirjalili2019,cao2018comprehensive} that favours a high inertial weight at the beginning of the search, which progressively drops over iterations: 
\begin{equation}
    w^k = w_{max} - \frac{w_{max}-w_{min}}{max\_iter} k
\end{equation}

\noindent which favours exploration at the beginning of the search and exploitation as the search progresses to the end to eventually constrict the search down to the area containing the best fitness and explore it in detail. The value of $w_{max}=0.9$ and $w_{min}=0.4$ are typically used in the literature.
Algorithm \ref{alg:pso} presents the typical pseudo code of the PSO search procedure.\newline

\begin{algorithm}[H]
\setstretch{0.6}
\SetAlgoLined
  Let $X_0=[x^l, x^u],V=[-v_{max},v_{max}]$\;
  Set swarm size N, max\_iter $K$ and $k=0$\;
  Randomly generate initial population: $x_i  \in X_0$\;
  Randomly generate initial velocities: $ v_i \in V$\;
  Set best $p_i^k$ for every particle to $x_i^k$\;
  Compute the swarm initial best point ($g^k$,BUB)\;
 \While{$k < K$ and \textbf{heuristic stop} not reached}{
  \For{i=1:N}{
    $v_i^{k+1}=\omega^kv_i^{k}+ v_1^k(p_i^k-x_i^k)+v_2^k(g^k-x_i^k)$\;
    $v_i^{k+1} = \text{bound(}v^{k+1},-v_{max},v_{max})$\;
    $x_i^{k+1} = x_i^k + v_i^{k+1}$\;
    $x_i^{k+1} = \text{bound(}x^{k+1},x^l,x^u)$\;
    $f_i^{k+1} = f(x_i^{k+1})$\;
    $p_i^k = \text{arg min}(\{f_i^{k+1},f_i^k\})$\;
    $g^k = \text{arg min}(\{f(p_i^k),$BUB$\})$\; 
  }
  $k=k+1$\;
 }
 \textbf{return} $(x^*=g^k,\text{BUB})$\;
 \caption{Typical PSO procedure}
 \label{alg:pso}
\end{algorithm}

\subsection{Neighbourhood topologies}
\begin{figure}[h]
    \centering    \includegraphics[scale=0.5]{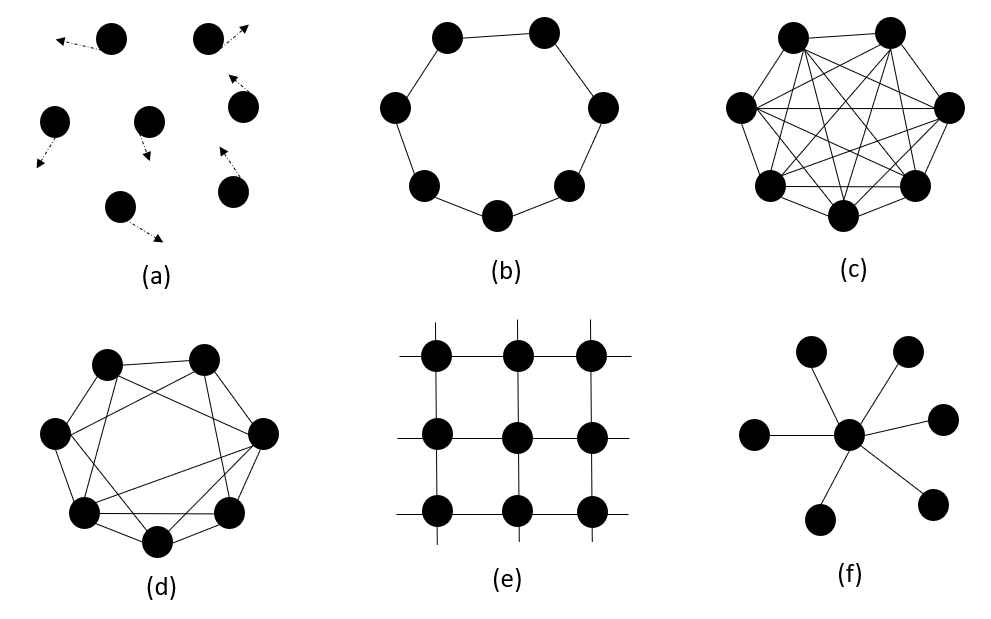}
    \caption{PSO neighbourhood topologies: (a) Cognitive model (b) lbest topology (c) gbest topology (d) small-world topology (e) Von Neumann topology (f) Wheel topology}
    \label{fig:topos}
\end{figure}

\noindent The gBest topology is the most widely known PSO organisation and the most used in unimodal global searches (Figure \ref{fig:topos}c). It is worth noting, however, that several other neighbourhood topologies exist that dictate how particles collaborate with each other, ranging from independent topologies, where particles are purely autonomous in their search, to collaborative topologies, where particles share information with each other.

\noindent 
The cognitive topology (See Figure \ref{fig:topos}a), for instance, is a topology where particles do not interact with each other, solely relying on cognitive experience. The ring topology is a topology in which particles interact with their adjacent neighbours based on their initial indices (See Figure \ref{fig:topos}b). The euclidean distance topology is a topology in which particles interact with their k neighbouring particles based on their proximity (See Figure \ref{fig:topos}b), the small-world topology is a topology whereby each particle's social information is obtained from a supervised random selection of particles in its neighbourhood, the Von Neumann topology (Figure \ref{fig:topos}e) where particles are organised in an edge-wrapped grid, such that each particle is connected to individuals above, below and to each side of it and the wheel topology in which particles connects to one focal particle to which they share information. These topologies share different attitudes centred around a trade-off between diversity and exploitation depending on the optimisation problem use case.

\section{Clustering-based multiswarm PSO algorithms}
\subsection{kPSO}
\noindent kPSO is a clustering-based multi-swarm PSO algorithm proposed by \cite{passaro2008particle}. The algorithm proceeds by a cyclic clustering of particles in the problem space to form independent swarms that exploit different promising regions independently. Initially, each particle proceeds autonomously using a cognitive motion model. At a regular interval $c$, k-means clustering is performed to join the closest particles together into niches that navigate using the classical gBest topology. To ensure diversity and the discovery of new regions, the worst particles in overcrowded niches are un-niched and permitted to navigate independently (i.e., cognitive model) until the next clustering cycle and the process repeats. Because k-means clustering requires an apriori knowledge of the number of clusters, \cite{passaro2008particle} proposes the use of the Bayesian information criterion (BIC) to adaptively estimate the number of clusters in the particles' formation. k-means clustering will run for k values between 2 to N/2, and the value with the highest BIC is selected as the number of clusters, thus, the number of niches. A c-value of 50 was recommended in \cite{passaro2008particle,passaro2007niching} as the iteration period for re-clustering. Algorithm \ref{alg:kpso} and \ref{alg:ikpso} describe the search mechanism of kPSO.

\begin{algorithm}[H]
\setstretch{0.6}
\SetAlgoLined
  Let $X_0=[x^l, x^u],V=[-v_{max},v_{max}]$\;
  Set swarm size N, max\_iter $K$ and $k=0$\;
  Randomly generate initial population: $x_i  \in X_0$\;
  Randomly generate initial velocities: $ v_i \in V$\;
  Set best $p_i^k$ for every particle to $x_i^k$\;
  Compute the swarm initial best point ($g^k$,BUB)\;
 \While{$k < K$ and \textbf{heuristic stop} not reached}{
  \If{k mod c=0}{
        Identify clusters\ $N_c$;
   }
   \For{j=1:$N_c$}{
      \For{i=1:$N_j$}{
        
        $v_{i,j}^{k+1}=\omega^kv_{i,j}^{k}+ c_1r_1(p_{i,j}^k-x_{i,j}^k) + c_2r_2(g_j^k-x_{i,j}^k)$\;
        $v_{i,j}^{k+1} = \text{bound(}v_{i,j}^{k+1},-v_{max},v_{max})$\;
        $x_{i,j}^{k+1} = x_{i,j}^k + v_{i,j}^{k+1}$\;
        $x_{i,j}^{k+1} = \text{bound(}x_{i,j}^{k+1},x^l,x^u)$\;
        $f_{i,j}^{k+1} = f(x_{i,j}^{k+1})$\;
        $p_{i,j}^k = \text{arg min}(\{f_{i,j}^{k+1},f_{i,j}^k\})$\;
        $g_j^k = \text{arg min}(\{f(p_{i,j}^k),$BUB$\})$\; 
      }
  }
  $k=k+1$\;
 }
 \textbf{return} $(x^*=g^k,\text{BUB})$\;
 \caption{Typical kPSO procedure}
 \label{alg:kpso}
\end{algorithm}
\begin{algorithm}[H]
\setstretch{0.6}
\SetAlgoLined
 Cluster particles' pbests using k-means\;
 Calculate the average number of particles per cluster, $N_{avg}$\;
 Set $N_u=0$\;

  \For{each cluster $C_j$}{
 \If{$N_j > N_{avg}$}{
        Remove the $N_j=N_{avg}$ worst particles from $C_j$\;
        Add $N_j-N_{avg}$ to $N_u$\;
 }
    Adapt the neighbourhood structure for the particles in $C_j$\;
 }
 Reinitialize the $N_u$ un-niched particles\;
\caption{Niche Identification procedure}
\label{alg:ikpso}
\end{algorithm}

\subsection{EDHC-PSO}

\noindent EDHC-PSO is another multiswarm clustering-based algorithm proposed by \cite{liu2020niching}. Unlike kPSO, EDHC-PSO does not perform cyclic clustering and only requires a single clustering pass. Upon initialisation of particles, each particle navigates in a semi-autonomous fashion using the euclidean-distance-based local best topology. When the cognitive experience of each particle stagnates after a number of iterations. Particles are clustered to form exploitation niches using the hierarchical clustering algorithm. Each created niche makes use of the small-world topology to improve the quality of the solution estimates in each niche. The algorithm was benchmarked against several other algorithms \cite{li2009niching,qu2012differential,qu2012distance,hui2015ensemble}, yielding the highest success rate. Nevertheless, the algorithms were tested on functions with typically a single global peak or not more than five global peaks, generally in a low dimensional space  (i.e., n=1,2). One major limitation of EDHC-PSO, however, is that it assumed an apriori number of clusters, unlike kPSO.
Algorithm \ref{alg:edhc-pso} describes the search mechanism of EDHC-PSO.

\begin{algorithm}[H]
\setstretch{0.6}
\SetAlgoLined
  Let $X_0=[x^l, x^u],V=[-v_{max},v_{max}]$\;
  Set swarm size N, max\_iter $K$ and $k=0$\;
  Generate the initial population: $x_i  \in X_0$\;
  Randomly generate initial velocities: $ v_i \in V$\;
  Set best $p_i^k$ for every particle to $x_i^k$\;
  
 \While{\textbf{all particles not stalled} and $iter < K$}{
  \For{i=1:N}{
       Move to next particle if $x_i$ has stalled\;
       Update and bound velocity $v_i^{k+1}$ using euclidean distance-lbest topology\;
       Update and bound the position $x_i^{k+1}$\;
       Compute the fitness of $x_i$ and update pbest\; 
  }
  $iter=iter+1$\;
 }
 Cluster particles using hierarchical clustering\;
\While{$iter < K$ and \textbf{all niche heads not stalled}}{
  \For{every niche $Nch$}{
  \For{i=1:Nch$_{size}$}{         
        Update and bound velocity $v_i^{k+1}$ using the small world topology model\;
        Update and bound the position $x_i^{k+1}$\;
        Compute the fitness of $x_i$ and update pbest\;
        Update the niche gbest if necessary\;
  }
  }
  $iter=iter+1$\;
 }
 \textbf{return} niche heads\;
 \caption{EDHC-PSO}
 \label{alg:edhc-pso}
\end{algorithm}

\subsection{NichePSO}
\noindent NichePSO is one of the first existing parallel niching PSO algorithms proposed by \cite{brits2002niching}. Rather than performing clustering per se, nichePSO makes use of merging operators to create new clusters. Initially, each particle navigates the problem space independently using the cognitive model topology. When the particle position does not sensibly vary upon a given number of iterations, a niche is formed that consists of the particle and its closest neighbour. A niche radius is formed between the best particle and the location of its furthest particle. Each niche makes use of the globally convergent PSO topology model proposed by \cite{van2002new}
to ensure that the gBest cannot stagnate when all vectors of the gBest coincide and the velocity is zero. When an independent particle enters the niche radius of an existing swarm, it is absorbed. Overlapping niches are potentially merged using a merging strategy.
\noindent The radius $R_j$ of a sub-swarm $S_j$ is defined as:
\begin{equation}
    R_j = max||x_g(j)-x||,x\in S_j
\end{equation}
\noindent where $g(j)$ is the best particle of the sub-swarm. A particle is absorbed into a subSwarm $S_j$ when it modes inside it:
\begin{equation}
    ||x-x_g(j)||\leq R_j
\end{equation}
\noindent and two subs-swarms $S_i$ and $S_j$ are merged when they intersect:
\begin{equation}
    ||x_g(j)-x_g(i)||<R_j+R_i
\end{equation}
In order for an independent particle in the main swarm to form a sub-swarm (i.e. partition criteria), its fitness value should show very little change after a number of iterations:
\begin{equation}
    |f_i^k-f_i^{k+m}|<\epsilon
\end{equation}
\noindent m is traditionally set to 3 \cite{brits2002niching}. NichePSO tends to merge several niche swarms into one when the merging criterion is too relaxed and thus can potentially lose good global optima. Several merging strategies have been proposed to mitigate the issue, the simplest being a no-merging approach \cite{engelbrecht2007enhancing,crane2020nichepso}. Algorithm \ref{alg:nichepso} describes the search mechanism of NichePSO.

\begin{algorithm}[h]
\setstretch{0.6}
\SetAlgoLined
  Let $X_0=[x^l, x^u], \; V=[-v_{\max}, v_{\max}]$\;
  Set swarm size $N$, max\_iter $K$ and $k=0$\;
  Randomly generate initial population: $x_i  \in X_0$\;
  Randomly generate initial velocities: $v_i \in V$\;
  Set best $p_i^k$ for every particle to $x_i^k$\;
  Compute the swarm initial best point ($g^k$, BUB)\;

 \While{$k < K$ and \textbf{heuristic stop} not reached}{
  \For{every $x_i \in$ mainSwarm}{
     $v_i^{k+1} = \omega^k v_i^{k} + v_1^k (p_i^k - x_i^k)$\;
     $x_i^{k+1} = \text{bound}(x_i^{k+1}, x^l, x^u)$\;
     $f_i^{k+1} = f(x_i^{k+1})$\;
     $p_i^k = \arg\min\{ f_i^{k+1},\, f_i^k \}$\;
     $g^k = \arg\min\{ f(p_i^k),\, \text{BUB} \}$\; 
  }

  \For{every subSwarm $S$}{
        Perform one step of the GCPSO algorithm \cite{van2002new}\; 
        Update subSwarm radius\;
  }  

  Possibly merge subSwarms\;

  \For{$i = 1:N$}{
    \eIf{$x_i \in$ mainSwarm \text{ and } $x_i$ meets partition criteria}{
       Create a new subSwarm with $x_i$\;
       Add its closest neighbour\;
    }{}
   }

  $k = k + 1$\;
 }

 \textbf{return} $(x^* = g^k, \text{BUB})$\;
 \caption{Niche PSO procedure}
 \label{alg:nichepso}
\end{algorithm}

\section{Proposed Enhancements} 
\noindent Prior to elaborating on the proposed enhancements, the current section first briefly explains the k-means clustering algorithm and the methodology used in this study to estimate the optimal number of clusters dynamically.

\subsection{k-means clustering}\label{sec:kmeans}
\noindent k-means clustering is a simple and efficient clustering algorithm that groups a set of n vectors given in k clusters centred of k centroids given an apriori number of clusters. The estimation of the optimal location of centroids is obtained by the minimisation of the within-cluster sum of square distance:

\begin{equation}
    WCSS = J(m_1,m_2,..,m_k) = \sum_{j=1}^{k}\sum_{j=1,x_j\in C_i}^{N_i}(x_j-m_i)^2
\end{equation}

\noindent where the $m_i$ vectors are the parametric centroid vectors, $N_i$ is the number of elements in cluster $C_i$. The objective function is generally minimised using Lyod's algorithm \cite{lloyd1982least}, which iteratively adapts initial centroid guesses. Alternatively, expensive metaheuristics are used to minimise the WCSS cost depending on the computational context \cite{kapil2016k}. To minimise the sensitivity of k-means clustering to initial guesses when using Lyod's algorithm, the algorithm is generally retrained several times, and the centroids of the hop with the least WCSS are selected.

\subsubsection{Estimating the number of clusters using Silhouette analysis}
\noindent Several techniques exist in the literature for the dynamic estimation of the number of clusters (i.e., elbow, gap statistic, Bayesian information criterion, canopy) \cite{j2020016,passaro2008particle}. In this study, the silhouette method \cite{dinh2019estimating} is used as a representative numerical criterion to estimate the number of clusters. The silhouette score estimates how poor or well is a data point dissimilarity within the assigned cluster than outside the cluster by computing two dissimilarity measures:
\begin{equation}
    a(x_i) = \frac{1}{N_p-1}\sum_{j,x_i\neq x_j\in C_p}^{N_p-1}d(x_i,x_j)
\end{equation}
\noindent where $a(x_i)$ is the average distance between a data point $x_i$ and all other points within its cluster $C_p$.
\begin{equation}
    b(x_i) = min\{\mu(x_i,C_j)\},j\neq p
\end{equation}
\noindent where $b(x_i)$ is the smallest average distance between a data point $x_i$ and any other point of all other clusters besides its cluster $C_p$.
The Silhouette score is thus denoted as

\begin{equation}
    s(x_i) = \frac{b(x_i)-a(x_i)}{max\{a(x_i),b(x_i)\}}, s(x_i) \in [-1,1]
\end{equation}
\noindent whereby the largest positive value indicates that the data point is well clustered and the largest negative value that it is poorly clustered. The average silhouette value  (i.e. $s_k=\frac{1}{N}\sum_{i}^{N}s(x_i)$) for all data points is used to rank a given k-Clustering.

\subsection{Contribution analysis of loosely connected topologies in CMPSO algorithms}
\label{sec:top_analysis}
\noindent The current study argues that neither the use of a cognitive model (equation \ref{eq:topos_cog}) or neighbourhood-centric topology (equations \ref{eq:topos_ec} and \ref{eq:topos_ring}) permits a sufficient exploration of the immediate surrounding of each particle. While the cognitive model provides an independent exploration of the particle neighbour, particles tend to revolve around their best experience thus far, which could hinder sufficient local exploration \cite{brits2002niching,kennedy1997particle}. On the other hand, the social component of neighbourhood-centric topology tends to favour the exploration of the best region among neighbours, also hindering independent local exploration, thus defeating the purpose of preliminary search in the CMPSO paradigm (i.e. maximal local information harvesting). 

\subsubsection*{Cognitive model:}
\begin{equation}
v_i^{k+1}=\omega^kv_i^{k}+ v_1^k(p_i^k-x_i^k)
\label{eq:topos_cog}
\end{equation}
\subsubsection*{Euclidean-based local neighbourhood:}
\begin{equation}
v_i^{k+1}=\omega^kv_i^{k}+ v_1^k(p_i^k-x_i^k)+v_2^k(g^k-x_i^k)
\label{eq:topos_ec}
\end{equation}
\subsubsection*{Ring topology:}
\begin{equation}
v_i^{k+1}=\omega^kv_i^{k}+ v_1^k(p_i^k-x_i^k)+v_2^k(g^k-x_i^k)
\label{eq:topos_ring}
\end{equation}

\noindent Therefore, a scouting phase around a bounded region of each particle is essential to get sufficient information about each particle surrounding and not miss out on any potential good region. This implies that for a given number of iterations and until the cognitive information is no longer enriched (i.e., $pBest$ improvement), it is relevant to explore the local region of each particle in the form 
\begin{equation}
    x_{k+1}^i = (x^i_u - x^i_l)e_k + x^i_l
\end{equation}

\noindent which can be reformulated as 
\begin{equation}
    x_{k+1} = x_{k} + v_{k+1}
\end{equation}

\begin{equation}
    v_{k+1} = x_0 - x_k - r + 2re_k
    \label{eq:scout_model}
\end{equation}

\noindent where $r$ is the radius of the hyper-sphere of search, $e_k$ is a uniformly distributed sequence range between 0 and 1. This sequence can be generated by any pattern generator that can produce even spread vectors across a bounded region. Candidate sequence generators are Halton-sequences, chaos and any other lowe discrepancy pseudo-number generator.
Assuming, a normalised problem space (i.e $|x_u| = |x_l|$), the value of r can be estimated as 
\begin{eqnarray}
    r =\gamma\frac{\sqrt{2}}{2}\sqrt[n]{\frac{\prod_{d=1}^{n}(x_d^u-x_d^l)}{N}}
    \label{eq:rad}
    % \\
    % d = arg\_max(x_d^u-x_d^l)
\end{eqnarray}

\noindent where $r$ is the scaled distance between the centre of a hypercube and one of its edges. The problem space is estimated as a set of N hypercubes, N being the population size. A $\gamma$ value of 1 is used in this study. Therefore a pervasive then cognitive model is proposed to sufficiently explore the search space and then further improve the resolution of the search in the particle region:
\begin{equation}
    v_{k+1} = 
    \begin{cases}
      x_0 - x_k - r + 2re_k\text{ when } |f(p_k)-f(p_{k+t})| \geq \epsilon_g \\
      \omega^kv_i^{k}+ v_1^k(p_i^k-x_i^k), \text{ otherwise}
    \end{cases}
    \label{eq:itopo}
\end{equation}

\noindent Halton sequence is used in this study for $e_k$ sequence generation due to its ability to generate evenly distributed data points in a bounded region with lower discrepancy than pseudo-random number generators \cite{weerasinghe2016particle}:
\begin{eqnarray}
    e(k) = (\phi_{p_1}(k),\phi_{p_2}(k),...,\phi_{p_d}(k))^T\\
    \phi_p(k) = \frac{b_0}{p}+\frac{b_1}{p^2}+...+\frac{b_m}{p^{m+1}}
\end{eqnarray}
\noindent where $p_1,p_2,..p_d$ are pairwise co-primes, $b_m$ are positive integers between 0 and p-1 forming the digits of each base $p$. For every $k$, a $\phi_{p_i}(k)$ number is formed at the selected $p$ base, which makes up a component of the $e_k$ vector.

\begin{figure}[h]
    \centering
    \includegraphics[scale=0.5]{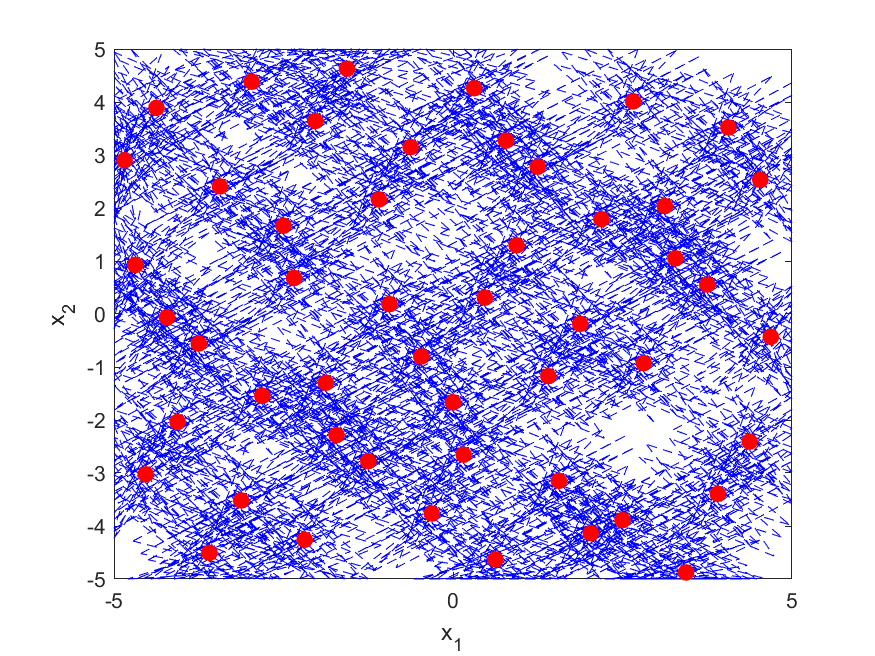}
    \caption{Halton-sequence based particle scouting (2D space)}
    \label{fig:my_label}
\end{figure}
\noindent A computational experiment is conducted in section \ref{sec:ce} to evaluate the performance of the proposed preliminary search topology against the cognitive model and the euclidean-distance model.

\subsection{The Clustering defect of CMPSO algorithms}
\label{sec:clust_def}
\begin{figure}[h]
    \centering
    \includegraphics[width=14cm]{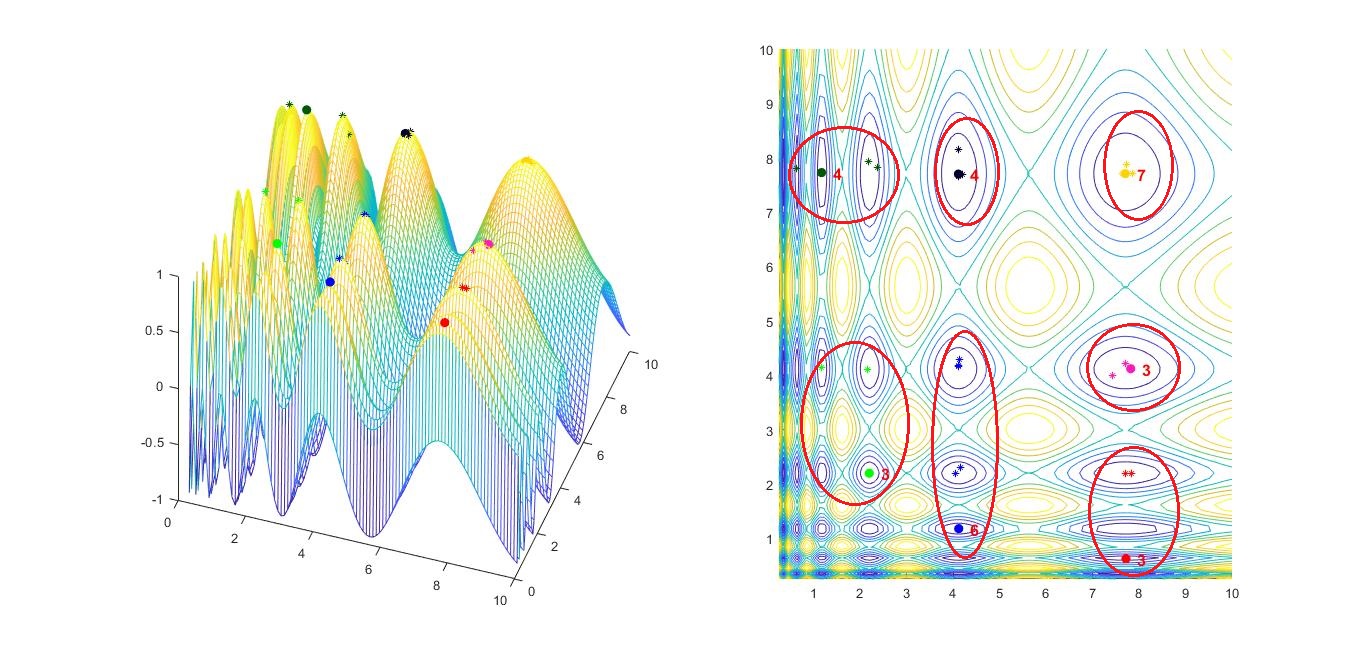}
    \caption{Distance-based Clustering Defect Illustration}
    \label{fig:cluster_defect}
\end{figure}

\noindent In CMPSO algorithms, the creation of clusters is immediately followed by the formation of niche swarms to exploit the estimated basins of convergence. It should be, however, noted that the newly created clusters may potentially contain more than one peak as the clustered particles do not necessarily belong to the same basin of convergence. In this way, using distance measures alone lacks resolution and disregards topological details of the problem space. Figure \ref{fig:cluster_defect} illustrates the problem. Assuming that potentially, the red ellipses in the 2D map of the problem space represent the clusters formed using distance features, the 3D map clearly shows that the clustered particles represent different peaks, and forming collaborative topologies on the estimated niches will lose some global optima. 

\noindent The current study proposes an additional stage of concavity analysis within the formed niches to investigate the presence of sub-niches. Concavity measure estimate can be used to estimate if two points belong to the same basin of convergence. The theory behind these measures rests on the \textit{convexity proposition}: 

\noindent $f(x)$, in $x \in [x^l,x^u]$ is convex if $\forall x_1,x_2 \in [x^l,x^u],x_1 \neq x_2: f(\lambda x_1 + (1-\lambda)x_2) \leq \lambda f(x_1) + (1-\lambda)f(x_2), \lambda  \in [0,1]$ which states that a function is convex in its bounded region provided that the line between any two-point pair is above the function. In the same vein, a function is concave if the line between any two-point is below the function. One such computational estimate is the Hill-valley algorithm proposed by \cite{ursem1999multinational}(See algorithm \ref{alg:concav_test1}), which can help estimate whether two points belong to the same peak or not.

\begin{algorithm}[H]
\setstretch{0.6}
\SetAlgoLined
Concavity\_Test($x_1,x_2,N$):\;
Let $X_{12}$, a set of $N$ points from the line between $x_1$ and $x_2$\;
$f_{min}$ = min($f(x_1),f(x_2)$)\;
\For{i=1:N}{
  select $x_i\in X_{12}$\;
  \If{$f(x_i) > f_{min}$)}{
    return 1\;
  }
}
return 0\;
\caption{Hill Valley concavity test algorithm}
 \label{alg:concav_test1}
\end{algorithm}

\noindent Interior points ($z$) within the segment that passes through $x_1$ and $x_2$ are obtained from equation \ref{eq:int_pt}.

\begin{equation}
   z = x_1 + \lambda(x_2-x_1), \lambda \in [0,1]
    \label{eq:int_pt}
\end{equation}
A typical sample size of 5 with $\lambda = [0.02,0.25,0.5,0.75,0.98]$ is used in the literature \cite{zhang2006novel}. Using the Hill-valley concavity measure, an elitist sub-clustering algorithm is proposed. 

\begin{algorithm}[H]
  Given a cluster $C_p$ of $q$ particles\;
  Optionally, perform local optimisation on the $q$ particles\;
  Select all $n$ particles where $f_p^*-f(x) < \epsilon$\;
  Create $n$ sub-clusters\;
  Sort sub-clusters from the fittest descendingly into set $S$\;
  \While{$S \neq \emptyset$}{
   Remove the least fit cluster head $g_i$\;
   \For{every subcluster head $g_j (j\neq i )$}{
       \If{$ hill\_valley\_func(g_i,g_j) = 0$}{
         merge cluster $SB_i$ to $SB_j$\;
         break\;
       }
   }
  }
  Obtain $m (m \leq n) $ sub-clusters after merging\;
  Compute the peak radius $r = min(||g_i-g_j||)/2$\;
  Allocate equitably $\frac{q}{m}-1$ exploitation particles around each peak within $r$ radius\;
  \caption{Sub-clustering algorithm}
  \label{alg:subcluster}
\end{algorithm}

\noindent Typically, given a cluster of $q$ particles, a pre-selection phase is performed where only the fittest particles within the cluster are used to minimise the creation of too many subclusters, therefore obtaining $n$ subclusters ($n \leq q$) initially. These subclusters of one particle each are sorted from the fittest to the least fit by function values loaded into a repository set $S$. Iteratively, the least subcluster is popped from the set $S$ and merges with a subcluster with a fitter cluster head if they belong to the same basin of convergence (i.e. hill\_valley\_func($g_i,g_j$) = 0), this process continues until the repository $S$ is emptied, therefor forming $m$ niches from a single cluster. To ensure that each sub-niche has enough particles to fine-tune the search, the least fit particles in sub-niches with more than $\frac{q}{m}$ particles are re-assigned to the other niches. The benefit of the sub-clustering is that when all particles within a cluster belong to the same basin convergence, the proposed algorithm performs conceptually with equal performance as conventional CMPSO algorithms. However, when several basins of convergence exist within a cluster, the proposed algorithm can identify sub-niches and thus has a higher peak ratio capacity than conventional CMPSO algorithms.

\noindent In addition, due to the subtlety of setting the threshold $\epsilon$ value to select initial sub-clusters and obtain consistent results, the current study proposes to first post-optimise the initial q clusters using local search methods and only to select the most qualitative particles (or potential basins) to form the niches. An SQP-based local search method can be used when the problem space is differentiable, and algorithms such as pattern search can be used when the problem space is derivative-free \cite{kramer2011derivative}.

\begin{figure}[h]
    \centering    \includegraphics[scale=0.7]{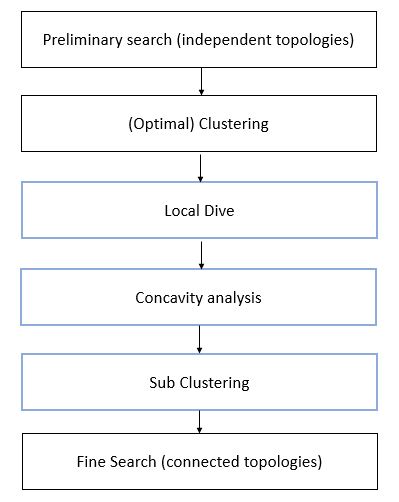}
    \caption{Proposed algorithm framework}
    \label{fig:my_label}
\end{figure}

\subsection{Proposed algorithm}
\noindent Using the two enhancements presented in section \ref{sec:top_analysis} and \ref{sec:clust_def}, an enhanced search framework is proposed (see Figure \ref{fig:cluster_defect}), and the overall algorithm is presented in the pseudo-code below (See algorithm \ref{alg:p-pso}). For the sake of brevity, the algorithm is coined as TImPSO, referred to as a topologically informed multi-swarm PSO.

\begin{algorithm}[H]
\setstretch{0.5}
\SetAlgoLined
  Let $X_0=[x^l, x^u],V=[-v_{max},v_{max}]$\;
  Set swarm size N, max\_iter $K$ and $k=0$\;
  Use Halton sequence to generate the initial population: $x_i  \in X_0$\;
  Randomly generate initial velocities: $ v_i \in V$\;
  Set best $p_i^k$ for every particle to $x_i^k$\;
  Compute the swarm initial best point ($g^k$,BUB)\;
 \While{\textbf{all particles not stalled} and $iter < K$}{
  \For{i=1:N}{   
       Move to next particle if $x_i$ has stalled\;
       Update and bound velocity $v_i^{k+1}$ using independent topology model (Eq \ref{eq:itopo})\;
       Update and bound the position $x_i^{k+1}$\;
       Compute the fitness of $x_i$ and update pbest\;        
  }
  $iter=iter+1$\;
 }
 Cluster the particles with an optimal number of clusters (see section \ref{sec:kmeans})\;
 \For{every \textbf{cluster}}{
    Perform sub-clustering analysis using Algorithm \ref{alg:subcluster} to form niches\;
 }

\While{$iter < K$ and \textbf{all niche heads not stalled}}{
  \For{every niche $Nch$}{
  \For{i=1:Nch$_{size}$}{          
       Update and bound velocity $v_i^{k+1}$ using collaborative topology model\;
       Update and bound the position $x_i^{k+1}$\;
       Compute the fitness of $x_i$ and update pbest\;
       Update the niche gbest if necessary\;    
  }
  }
  $iter=iter+1$\;
 }
 \textbf{return} niche heads\;
 \caption{TImPSO}
 \label{alg:p-pso}
\end{algorithm}

\begin{table}[h]
    \centering
    \begin{tabular}{|c|c|}
         \hline
         Preliminary search &  Pervasive-cognitive model\\
         \hline
         Clustering & k-means using silhouette method\\
         \hline
         SubClustering & Elitist Ursem-based Concavity analysis\\
         \hline
         Fine-search & Equitable particle allocation + gBest 
 PSO\\
         % \hline
         % Scalability & Sequential niching using hashing-based mapping\\
         \hline
    \end{tabular}    \caption{Characteristics summary of the TImPSO algorithm}
    \label{tab:my_label}
\end{table}

\section{Computational experiments}\label{sec:ce}
\subsection{Performance measure}
\noindent The performance of the proposed method has been tested on IEEE CEC2013 test functions designed to benchmark niching algorithms \cite{li2013benchmark}. The test sets comprise twenty multimodal test functions of multiple structures ranging from 1D to 20D with several sparsely distributed global optima. (Table \ref{tab:test_funcs}).  The peak ratio of each algorithm across all test functions has been recorded.
\subsubsection*{Peak Ratio}
\noindent Given a maximum number of function evaluations and a required accuracy level, the Peak Ratio (PR) measures the average percentage of all known global optima found over multiple runs:
\begin{equation}
    PR = \frac{\sum_{i=1}^{N_r}N_{gf}}{N_g * N_r}
\end{equation}
\noindent where $N_{gf}$ denotes the number of global optima found, $N_g$, the number of global optima existing in the function and $N_r$, the number of optimisation runs.

\subsection{Initialisation and search settings}
\noindent To ensure that all particles are equally distributed across the problem space at the beginning of the search, Halton sequence was used for population initialisation \cite{weerasinghe2016particle}. 
Three computational experiments were conducted. First, an experiment was conducted to assess the performance of preliminary search topologies. The euclidean-distance-based topology (NPSOe), the cognitive topology (NPSOc) and the proposed pervasive-then-cognitive topology (NPSOhc) were used interchangeably for the preliminary phase, followed by k-means clustering using the silhouette method to estimate the optimal number of clusters. This experiment aimed to evaluate which model is more likely to aid in discovering basins of convergence among the three topologies. Table \ref{tab:sim-prelim} presents records of peak performances for each topology. The experiment was conducted for 30 optimisation runs, and the average peak ratios for each function across five accuracy levels ($ 10^{-1},10^{-2},10^{-3},10^{-4},10^{-5}$) were recorded in accordance with CEC niching benchmarking recommendations \cite{li2013benchmark}. The peak ratio was recorded based on the number of niches formed and based on the accuracy level of each niche gBest.

\noindent Secondly, an experiment was conducted comparing the performance of the proposed enhanced algorithm (TImPSO) against competitive algorithms: NichePSO \citep{brits2002niching}, kPSO \citep{passaro2008particle} and EDHC-PSO \citep{liu2020niching}. The peak ratio performances of each algorithm were recorded and tested on the CEC2013 datasets in the same setting as the first experiment. k-Means clustering with silhouette evaluation was used in the same setting for $k$PSO, EDHC-PSO and TImPSO for a fair comparison of the performance of the algorithms. The gBest topology was used as the collaborative model for all EDHC-PSO, $k$PSO and TImPSO. A modified EDHC-PSO algorithm is used utilising the proposed preliminary search due to the poor performance of the euclidean-based lbest topology in multiple peak settings (See Table \ref{table:perf_timpso1}). A No merging approach was used for NichePSO to ensure that no formed niches are potentially absorbed. The SQP local search was used for the local dives of cluster particles in TImPSO since all test functions were differentiable. A thresholding epsilon of 0.1 was used to pre-select particles that are eligible for concavity analysis.

\noindent A third experiment was conducted whereby the results (i.e. niche head particles) of all four algorithms were post-optimised using SQP local searches to assess how performant was TImPSO if all algorithms niche heads values were post-optimised, more importantly for NichePSO \cite{crane2020nichepso}. The average peak ratios of all test functions were recorded in the same settings as in previous experiments. A population size of 30 was used across all test functions, and all algorithms ran for 1000 function evaluations. The Wilcoxon sign rank test was used to assess how statistically different the median peak ratio per function in each conventional method was compared to the proposed enhancements. Table \ref{tab:conf1} and \ref{tab:conf2} summarise the configurations of the experiments.
\begin{table}[h]
    \centering
    \scalebox{0.7}{
    \begin{tabular}{|c|l|}
        \hline
         NPSOhc &\begin{tabular}{l}preliminary-phase: halton+cognitive ($w=0.7290$)\\ clustering: kMeans + silhouette\\fine-search phase: gBest topos ($w_{max}=0.9, w_{min}=0.4,s_r=c_r=2$)  \end{tabular}\\         
         \hline
         NPSOe &\begin{tabular}{l}preliminary-phase: euclidean lBest ($w=0.7290$, neighbourhood size=1)\\ clustering: kMeans + silhouette\\fine-search phase: gBest topos ($w_{max}=0.9, w_{min}=0.4,s_r=c_r=2$)  \end{tabular}\\
         \hline
         NPSOc &\begin{tabular}{l}preliminary-phase: cognitive ($w=0.7290$)\\ clustering: kMeans + silhouette\\fine-search phase: gBest topos ($w_{max}=0.9, w_{min}=0.4,s_r=c_r=2$) \end{tabular}\\      
         \hline         
    \end{tabular}
    }
    \caption{Parameter configuration of test algorithms for experiment 1. Population size = 30}
    \label{tab:conf1}
\end{table}

\begin{table}[h]
    \centering
    \scalebox{0.7}{
    \begin{tabular}{|c|l|}
        \hline
         EDHC-PSO &\begin{tabular}{l}preliminary-phase: halton+cognitive ($w=0.7290$)\\ clustering: kMeans + silhouette\\fine-search phase: gBest toposgBest topos 
 ($w_{max}=0.9, w_{min}=0.4,s_r=c_r=2$) \end{tabular}\\         
         \hline
         NichePSO &\begin{tabular}{l}topology: cognitive ($w=0.7290$)\\ subswarm merging strategy: No Merge\\sub-swarm topology: GCPSO ($w=0.7290,s_r=c_r=2$) \end{tabular}\\
         \hline
         kPSO &\begin{tabular}{l}preliminary-phase: halton+cognitive ($w=0.7290$)\\ clustering: kMeans + silhouette
         \\ clustering iteration cycle ($c$): 50\\
         fine-search phase: gBest topos ($w_{max}=0.9, w_{min}=0.4,s_r=c_r=2$) \end{tabular}\\
         \hline
         TImPSO &\begin{tabular}{l}preliminary-phase: halton+cognitive ($w=0.7290$)\\ clustering: kMeans + silhouette
         \\ sub-clustering: Ursem-based concavity analysis\\
         fine-search phase: gBest topos ($w_{max}=0.9, w_{min}=0.4,s_r=c_r=2$) \end{tabular}\\         
         \hline         
    \end{tabular}
    }
    \caption{Parameter configuration of test algorithms for experiments 2 and 3. In experiment 3, every niche head is fine-tuned using a local SQP search. Population size = 30}
    \label{tab:conf2}
\end{table}

\begin{table}[H]
      \centering   \scalebox{0.75}{
     \begin{tabular}{|c|l|c|c|}
     \hline
           $f_n$ & Functions & Range &  No. global optima \\
           \hline           \multicolumn{4}{|c|}{CEC 2013}\\
          \hline
          $f_1$& Five-Uneven-Peak Trap - 1D & [0,30] & 2\\
           $f_2$& Equal Maxima - 1D & [0,1] & 5\\
           $f_3$& Uneven Decreasing Maxima - 1D & [0,1] & 1\\
          $f_4$& Himmelblau - 2D  & [-6,6]$^D$ & 4\\
           $f_5$& Six-Hump Camel Back 2D & [-1.9,1.9] [-1.1 1.1] & 2\\
          $f_{6}$& Shubert - 2D  & [-10,10]$^D$ & 18\\
          $f_{7}$& Shubert - 3D & [-10,10]$^D$ & 81\\
          $f_{8}$& Vincent - 2D & [0.25,10]$^D$ & 36\\
          $f_{9}$& Vincent - 3D & [0.25,10]$^D$ & 216\\
          $f_{10}$& Modified Rastrigin - 2D & [0,1]$^D$ & 12 \\
          $f_{11}$& Composition Function 1 - 2D & [-5,5]$^D$ & 6\\
          $f_{12}$& Composition Function 2 - 2D & [-5,5]$^D$ & 8\\
          $f_{13}$& Composition Function 3 - 2D & [-5,5]$^D$ & 6\\
          $f_{14}$& Composition Function 3 - 3D & [-5,5]$^D$ & 6\\
          $f_{15}$& Composition Function 3 - 5D & [-5,5]$^D$ & 6\\
          $f_{16}$& Composition Function 3 - 10D & [-5,5]$^D$ & 6\\
          $f_{17}$& Composition Function 4 - 3D & [-5,5]$^D$ & 8\\
          $f_{18}$& Composition Function 4 - 5D & [-5,5]$^D$ & 8\\
          $f_{19}$& Composition Function 4 - 10D & [-5,5]$^D$ & 8\\
          $f_{20}$& Composition Function 4 - 20D & [-5,5]$^D$ & 8\\
         \hline
     \end{tabular}
     }
     \caption{CEC 2013 Special session niche test functions \cite{li2013benchmark}}
     \label{tab:test_funcs}
 \end{table}

\section{Results and Discussion}
\subsection{On the performance of the preliminary search topologies}
\noindent Table \ref{tab:sim-prelim} presents the performance of the two commonly used search topologies in CMPSO algorithms. The results in the table narrate that the proposed topology that consists of scouting first the problem space prior to independent particle search has a higher peak ratio compared to a purely cognitive model and is much superior to the euclidean-distance-based approach. These findings fall in line with the initial hypothesis that an extensive exploration increases the likelihood of finding the most promising regions and better inform clustering analysis. As advertised in section \ref{sec:top_analysis}, other ergodic distribution approaches could also be attempted in place of the Halton sequence, such as chaos \cite{tian2017particle} and spiral dynamics \cite{tamura2017spiral}. Nevertheless, the cognitive model as remains a competitive preliminary search neighbourhood topology which the current proposed model further improves. In light of the current experiment, the euclidean-distanced based lbest model is not a competitive preliminary search approach with poor concurrent exploration. The model has only been competitive in lower dimensional space for problems with few peaks (i.e. $<$ 3).

\begin{table}[h]                                 
\centering                     
% \begin{tabular}{>{\centering\arraybackslash}p{2 cm}p{2 cm}p{2 cm}p{2 cm}}   
\scalebox{0.95}{
\begin{tabular}{cccc}   
 \hline
 Func.& NPSOhc & NPSOe & NPSOc \\                
 \hline
$f_{1}$ & \textbf{1.00$\pm$0.000} & \textbf{1.00$\pm$0.000$\approx$} & \textbf{1.00$\pm$0.000$\approx$} \\  
$f_{2}$ & \textbf{0.96$\pm$0.081} & 0.29$\pm$0.104- &\textbf{ 0.99$\pm$0.037$\approx$} \\ 
$f_{3}$ & 1.00$\pm$0.061 & \textbf{1.00$\pm$0.069$\approx$} & \textbf{1.00$\pm$0.000$\approx$} \\  
$f_{4}$ &\textbf{ 1.00$\pm$0.000 }& 0.28$\pm$0.132- & \textbf{0.98$\pm$0.063$\approx$} \\ 
$f_{5}$ & \textbf{1.00$\pm$0.000} & \textbf{1.00$\pm$0.000$\approx$} & \textbf{1.00$\pm$0.000$\approx$} \\  
$f_{6}$ &\textbf{ 0.34$\pm$0.033 }& 0.00$\pm$0.003- & 0.24$\pm$0.040- \\         
$f_{7}$ &\textbf{ 0.05$\pm$0.012 }& 0.00$\pm$0.000- & 0.04$\pm$0.013- \\         
$f_{8}$ &\textbf{ 0.22$\pm$0.016 }& 0.08$\pm$0.016- & 0.18$\pm$0.051- \\         
$f_{9}$ &\textbf{ 0.04$\pm$0.005 }& 0.01$\pm$0.002- & 0.03$\pm$0.006- \\         
$f_{10}$ &\textbf{ 0.80$\pm$0.063 }& 0.19$\pm$0.035- & 0.71$\pm$0.109- \\        
$f_{11}$ & 0.51$\pm$0.030 & 0.40$\pm$0.089- &\textbf{ 0.70$\pm$0.140+} \\        
$f_{12}$ &\textbf{ 0.67$\pm$0.219 }& 0.07$\pm$0.043- & 0.39$\pm$0.133- \\        
$f_{13}$ &\textbf{ 0.62$\pm$0.090 }& 0.22$\pm$0.133- & 0.57$\pm$0.093- \\        
$f_{14}$ &\textbf{ 0.67$\pm$0.006 }& 0.15$\pm$0.079- & 0.45$\pm$0.158- \\        
$f_{15}$ & 0.13$\pm$0.113 & 0.00$\pm$0.006- &\textbf{ 0.15$\pm$0.110$\approx$} \\
$f_{16}$ &\textbf{ 0.02$\pm$0.047 }& 0.00$\pm$0.000- & \textbf{0.01$\pm$0.035$\approx$} \\
$f_{17}$ &\textbf{ 0.42$\pm$0.074 }& 0.02$\pm$0.039- & 0.19$\pm$0.102- \\        
$f_{18}$ &\textbf{ 0.12$\pm$0.036 }& 0.00$\pm$0.000- & 0.05$\pm$0.068- \\        
$f_{19}$ & 0.00$\pm$0.000 & 0.00$\pm$0.000$\approx$ & 0.00$\pm$0.000$\approx$ \\ 
$f_{20}$ & 0.00$\pm$0.000 & 0.00$\pm$0.000$\approx$ & 0.00$\pm$0.000$\approx$ \\
\hline
\end{tabular} 
}
\caption{Analysis of peak ratio performance of independent topologies. Average of 30 optimisation runs. The $-$, $+$, $\approx$ signs indicate respectively that the solver was worst, better or similar compared to TImPSO with statistical significance according to the Wilcoxon sign rank test ($\alpha=0.05$).}                 
\label{tab:sim-prelim}          
\end{table} 

\subsection{On the performance of the augmented clustering algorithm}
\noindent Table \ref{table:perf_timpso1} shows the average peak ratio of all four algorithms on the CEC 2013 testbench. The results show that TImPSO has a superior peak identification ratio compared to $k$PSO, EDHC-PSO and NichePSO, obtaining the highest peak ratio in nineteen problems out of twenty. Also, when the niche heads in each algorithm are further fine-tuned by SQP local search, the proposed algorithm still yields the highest peak ratio among all algorithms asserting the benefit of the proposed enhancements. These resullts demonstrate empirically how using proximity features alone during clustering leads to the loss of good peaks within the formed niches. Thus the inclusion of geometric topology analysis is relevant in enhancing the information depth of the optimisation algorithm and thus helps devise better optimisation decisions well aware of the problem landscape. Further investigation should, however, be done to test the performance and robustness of the proposed algorithm (TImPSO) in practical optimisation problems with additional adjustments where necessary.

\begin{table}[H]         
\centering      
\scalebox{0.95}{        
\begin{tabular}{ccccc}   
\hline
 & kPSO & EDHC-PSO & NichePSO & TimPSO \\       
\hline
$f_{1}$ & 0.73$\pm$0.254- &\textbf{ 1.00$\pm$0.000$\approx$ }& \textbf{1.00$\pm$0.000$\approx$} & \textbf{1.00$\pm$0.000} \\         
$f_{2}$ &\textbf{ 0.99$\pm$0.037$\approx$ }& 0.49$\pm$0.172- & 0.39$\pm$0.128- & \textbf{0.99$\pm$0.051}\\                 
$f_{3}$ & \textbf{1.00$\pm$0.000$\approx$} & \textbf{1.00$\pm$0.081$\approx$} & \textbf{1.00$\pm$0.081$\approx$} & \textbf{1.00$\pm$0.069} \\          
$f_{4}$ & 0.90$\pm$0.193- & \textbf{0.98$\pm$0.063$\approx$} & 0.93$\pm$0.123- &\textbf{ 1.00$\pm$0.000} \\                 
$f_{5}$ & 0.94$\pm$0.110- &\textbf{ 1.00$\pm$0.000$\approx$ }& \textbf{0.98$\pm$0.091$\approx$} & \textbf{1.00$\pm$0.000} \\         
$f_{6}$ & 0.32$\pm$0.048- & 0.23$\pm$0.060- & 0.22$\pm$0.054- &\textbf{ 0.55$\pm$0.051} \\                         
$f_{7}$ & 0.08$\pm$0.015- & 0.04$\pm$0.019- & 0.04$\pm$0.015- &\textbf{ 0.12$\pm$0.020} \\                         
$f_{8}$ & 0.20$\pm$0.035- & 0.18$\pm$0.033- & 0.19$\pm$0.030- &\textbf{ 0.40$\pm$0.020} \\                         
$f_{9}$ & 0.04$\pm$0.007- & 0.04$\pm$0.006- & 0.04$\pm$0.005- &\textbf{ 0.10$\pm$0.006} \\                         
$f_{10}$ & 0.72$\pm$0.158- & 0.50$\pm$0.092- & 0.52$\pm$0.110- &\textbf{ 1.00$\pm$0.000} \\                        
$f_{11}$ & 0.74$\pm$0.203- & 0.74$\pm$0.113- & 0.70$\pm$0.113- &\textbf{ 0.84$\pm$0.133} \\                        
$f_{12}$ & 0.43$\pm$0.211- & 0.47$\pm$0.127- & 0.49$\pm$0.098- &\textbf{ 0.80$\pm$0.089} \\                        
$f_{13}$ & 0.61$\pm$0.173- & 0.56$\pm$0.113- & 0.60$\pm$0.102- &\textbf{ 0.72$\pm$0.079} \\                        
$f_{14}$ & 0.52$\pm$0.230- & 0.55$\pm$0.107- & 0.53$\pm$0.100- &\textbf{ 0.67$\pm$0.018} \\                        
$f_{15}$ & 0.14$\pm$0.116- & 0.24$\pm$0.142- & 0.20$\pm$0.120- &\textbf{ 0.38$\pm$0.109} \\                        
$f_{16}$ & 0.01$\pm$0.031- & 0.00$\pm$0.013- & 0.00$\pm$0.000- &\textbf{ 0.11$\pm$0.101} \\                        
$f_{17}$ & 0.23$\pm$0.163- & 0.25$\pm$0.122- & 0.33$\pm$0.116- &\textbf{ 0.53$\pm$0.100} \\                  
$f_{18}$ & 0.10$\pm$0.075- & 0.09$\pm$0.082- & 0.11$\pm$0.070- &\textbf{ 0.18$\pm$0.071} \\                   
$f_{19}$ &\textbf{ 0.01$\pm$0.029$\approx$ }& 0.00$\pm$0.000$\approx$ & 0.00$\pm$0.000$\approx$ & \textbf{0.01$\pm$0.023}\\
$f_{20}$ & 0.00$\pm$0.000$\approx$ & 0.00$\pm$0.000$\approx$ & 0.00$\pm$0.000$\approx$ & 0.00$\pm$0.000 \\        
\hline
\end{tabular}  
}                         
\caption{Niching performance (Average of 30 optimisation runs). The $-$, $+$, $\approx$ signs indicate respectively that the solver was worst, better or similar compared to TImPSO with statistical significance according to the Wilcoxon sign rank test ($\alpha=0.05$).}           
\label{table:perf_timpso1}                       
\end{table} 

\begin{table}[H]         
\centering              
\begin{tabular}{ccccc}   
\hline
 & kPSO & EDHCPSO & NichePSO& TImPSO \\  
\hline  
$f_{1}$ & 0.67$\pm$0.240- &\textbf{ 1.00$\pm$0.000$\approx$ }& \textbf{1.00$\pm$0.000$\approx$} & \textbf{1.00$\pm$0.000} \\        
$f_{2}$ &\textbf{ 0.99$\pm$0.073$\approx$ }& 0.49$\pm$0.201- & 0.41$\pm$0.170- & \textbf{0.97$\pm$0.076} \\                
$f_{3}$ & \textbf{1.00$\pm$0.037$\approx$} & \textbf{1.00$\pm$0.086$\approx$} & \textbf{1.00$\pm$0.076$\approx$} & \textbf{1.00$\pm$0.037} \\         
$f_{4}$ & \textbf{0.96$\pm$0.133$\approx$} & \textbf{0.98$\pm$0.063$\approx$} & \textbf{0.96$\pm$0.095$\approx$} &\textbf{ 1.00$\pm$0.000} \\
$f_{5}$ &\textbf{ 1.00$\pm$0.000$\approx$ }& \textbf{0.98$\pm$0.091$\approx$} & \textbf{1.00$\pm$0.000$\approx$} & \textbf{1.00$\pm$0.000} \\
$f_{6}$ & 0.33$\pm$0.060- & 0.25$\pm$0.048- & 0.23$\pm$0.042- &\textbf{ 0.56$\pm$0.064} \\                        
$f_{7}$ & 0.09$\pm$0.023- & 0.05$\pm$0.016- & 0.05$\pm$0.014- &\textbf{ 0.13$\pm$0.022} \\                        
$f_{8}$ & 0.21$\pm$0.054- & 0.20$\pm$0.042- & 0.19$\pm$0.028- &\textbf{ 0.40$\pm$0.016} \\                        
$f_{9}$ & 0.05$\pm$0.007- & 0.04$\pm$0.008- & 0.04$\pm$0.007- &\textbf{ 0.10$\pm$0.006} \\                        
$f_{10}$ & 0.81$\pm$0.097- & 0.52$\pm$0.104- & 0.55$\pm$0.125- &\textbf{ 1.00$\pm$0.000} \\                       
$f_{11}$ & 0.78$\pm$0.184- & 0.71$\pm$0.109- & 0.72$\pm$0.104- &\textbf{ 0.88$\pm$0.137} \\                       
$f_{12}$ & 0.50$\pm$0.141- & 0.49$\pm$0.123- & 0.50$\pm$0.150- &\textbf{ 0.84$\pm$0.099} \\                       
$f_{13}$ & \textbf{0.69$\pm$0.124$\approx$} & 0.55$\pm$0.117- & 0.58$\pm$0.122- &\textbf{ 0.71$\pm$0.083} \\               
$f_{14}$ & 0.55$\pm$0.199- & 0.54$\pm$0.136- & 0.52$\pm$0.129- &\textbf{ 0.67$\pm$0.000} \\                       
$f_{15}$ & 0.17$\pm$0.107- & 0.22$\pm$0.147- & 0.20$\pm$0.131- &\textbf{ 0.38$\pm$0.153} \\                       
$f_{16}$ & 0.02$\pm$0.058- & 0.00$\pm$0.019- & 0.01$\pm$0.042- &\textbf{ 0.09$\pm$0.084} \\                       
$f_{17}$ & 0.30$\pm$0.101- & 0.32$\pm$0.122- & 0.36$\pm$0.111- &\textbf{ 0.50$\pm$0.111} \\                       
$f_{18}$ & 0.10$\pm$0.089- & 0.09$\pm$0.077- & 0.07$\pm$0.058- &\textbf{ 0.17$\pm$0.068} \\                       
$f_{19}$ & 0.00$\pm$0.000$\approx$ & 0.00$\pm$0.000$\approx$ & 0.00$\pm$0.000$\approx$ & 0.00$\pm$0.000 \\        
$f_{20}$ & 0.00$\pm$0.000$\approx$ & 0.00$\pm$0.000$\approx$ & 0.00$\pm$0.000$\approx$ & 0.00$\pm$0.000 \\
\hline
\end{tabular}                           
\caption{Niching performance with SQP post optimisation (Average of 30 optimisation runs). The $-$, $+$, $\approx$ signs indicate respectively that the solver was worst, better or similar compared to TImPSO with statistical significance according to the Wilcoxon sign rank test ($\alpha=0.05$).}                       \label{table:perf_timpso2}       
\end{table} 

\subsection{On limitations of parallel niching frameworks: Towards scalable algorithms}
\noindent The results in Tables \ref{table:perf_timpso1} and \ref{table:perf_timpso2} show how all the CMPSO algorithms do not attain 100\% success rate for functions with a large number of optima. This limitation is due to several reasons, including the number of particles used in the experiment but, more importantly, the limited peak identification capacity of parallel niching algorithms. First, it is not possible to know apriori the number of existing global optima in a given problem space; thus, a parallel niching algorithm cannot obtain more peaks than its population size. To mitigate this problem, the current study suggests that sequential niching is integrated to parallel niching. Although parallel niching is deemed faster in identifying niches concurrently, a restart of the algorithm should be performed while banning previously discovered regions. This way, a novel hybrid framework will be created that maintains the niche identification speed of the parallel niching algorithm in conjunction with the scalability of sequential niching \citep{beasley1993sequential,matanga2022frac,matanga2022nocp}. Future investigation will be invested in developing a scalable TImPSO algorithm that efficiently embeds sequential niching.

\section{Conclusion}
\noindent The current research has investigated the clustering-based multiswarm PSO paradigm pinpointing two limitations: insufficient exploration during the preliminary search phase and the low resolution of proximity-based clustering analysis. It has thus suggested an initial scouting phase in the neighbourhood of each initialised particle prior to undertaking an independent search using the cognitive model. This ensures that each particle provides sufficient local information about its region in a near-exhaustive approach to better inform clustering and not miss out on good promising areas. Also, the research has suggested a sub-clustering phase using concavity analysis that aims to investigate the existence of sub-niches within formed clusters. This has translated into an increased peak detection ratio, thus demonstrating the importance of geometry topology analysis within the CMPSO paradigm. These enhancements have been tested on IEEE CEC 2013 niching test sets yielding an improved peak ratio from conventional CMPSO algorithms. Further research will be directed towards embedding sequential niching within the parallel search framework to improve scalability when the number of existing peaks is large. Also, the proposed algorithm will be tested on practical problems with practical adjustments.

\section*{Acknowledgements}
\noindent This research was supported by South African National Research Foundation Grants (Nos. 132159, 114911, 137951 and 132797) and Tertiary Education Support Programme (TESP) of South African ESKOM.

% \bibliography{mybibfile}
% \section*{Bibliography}

\end{document}